\documentclass[10pt,twocolumn,letterpaper]{article}

\usepackage{cvpr}              

\usepackage{amsmath}
\usepackage{amssymb}
\usepackage{booktabs}
\usepackage{tabularx}

\usepackage{color}
\usepackage{multirow}
\usepackage{bbm}
\usepackage{rotating}
\usepackage{graphicx}

\makeatletter
\newcommand{\myline}{
    \noalign {\ifnum 0=`}\fi \hrule height 1pt
    \futurelet \reserved@a \@xhline
}
\newcolumntype{"}{@{\hskip \tabcolsep \vrule width 1pt \hskip \tabcolsep}}
\makeatother

\usepackage[dvipsnames]{xcolor}

\definecolor{cvprblue}{rgb}{0.21,0.49,0.74}
\usepackage[pagebackref,breaklinks,colorlinks,citecolor=cvprblue]{hyperref}

\def\paperID{6410} 
\def\confName{CVPR}
\def\confYear{2024}

\title{Frequency Domain Modality-invariant Feature Learning for Visible-infrared Person Re-Identification}

\author{Yulin Li$^1$, Tianzhu Zhang$^{1, }$\thanks{Corresponding Author},  Yongdong Zhang$^1$ \\
$^1$ University of Science and Technology of China\\
{\tt\small liyulin@mail.ustc.edu.cn~~~\{tzzhang, zhyd73\}@ustc.edu.cn}
}

\begin{document}
\maketitle

\begin{abstract}
Visible-infrared person re-identification (VI-ReID) is challenging due to the significant cross-modality discrepancies between visible and infrared images.
 While existing methods have focused on designing complex network architectures or using metric learning constraints to learn modality-invariant features, they often overlook which specific component of the image causes the modality discrepancy problem.
In this paper, we first reveal that the difference in the amplitude component of visible and infrared images is the primary factor that causes the modality discrepancy and further propose a novel Frequency Domain modality-invariant feature learning framework (FDMNet) to reduce modality discrepancy from the frequency domain perspective.
Our framework introduces two novel modules, namely the Instance-Adaptive Amplitude Filter (IAF) module and the Phrase-Preserving Normalization (PPNorm) module, to enhance the modality-invariant amplitude component and suppress the modality-specific component at both the image- and feature-levels.
Extensive experimental results on two standard benchmarks, SYSU-MM01 and RegDB, demonstrate the superior performance of our FDMNet against state-of-the-art methods.
\end{abstract}

\section{Introduction}
\label{sec:intro}
Person re-identification (ReID) aims to match person images captured by non-overlapping camera views~\cite{zheng2016person,ye2021deep}.
Due to its substantial practical value in intelligent surveillance systems, ReID has attracted substantial attention in the computer vision area.
Existing person ReID methods~\cite{hermans2017defense,zhong2017re,sun2018beyond,li2018harmonious,li2021diverse,he2021transreid,zhu2022dual} mainly focus on single-modality image matching, i.e., using images captured by visible cameras during the daytime.
However, these methods may perform poorly when visible cameras cannot capture valid appearance characteristics of a person under poor illumination environments, such as at night.
In modern surveillance systems, cameras often switch from visible to infrared mode when the light condition is undesirable.
Therefore, visible-infrared person re-identification (VI-ReID)~\cite{wu2017rgb} has recently become an essential task, which aims to match person images captured by visible and infrared cameras.

\begin{figure}[t]
\centering
\includegraphics[
width=0.99\columnwidth]{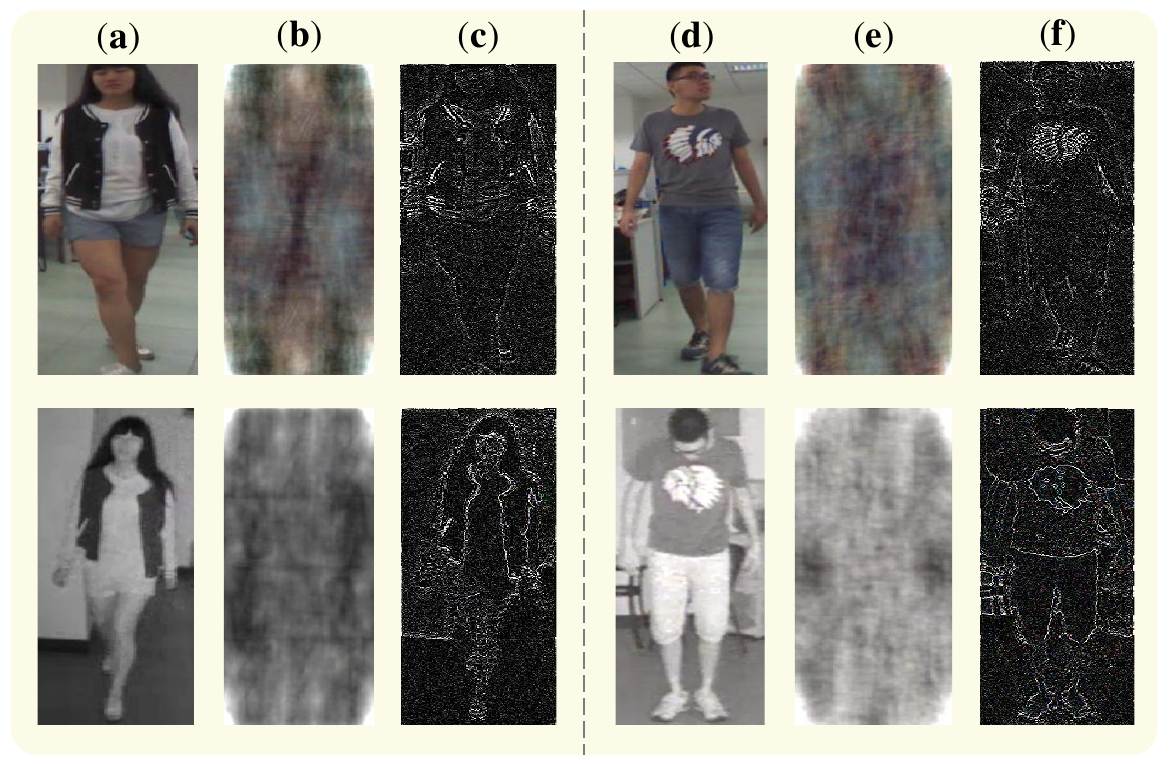}
\caption{
Examples of the amplitude-only and phase-only reconstruction.
(a) (d) Original images. (b) (e) Reconstructed images with amplitude information only by setting the phase component to a constant. (c) (f) Reconstructed images with phase information only by setting the amplitude component to a constant.
}
\label{motivation1}
\vspace{-6mm}
\end{figure}

Compared with the single-modality ReID task, VI-ReID encounters an additional modality discrepancy problem resulting from the different wavelength ranges used in the imaging process~\cite{sarfraz2017deep}.
Existing methods mainly alleviate the modality discrepancy from two aspects: image-level alignment~\cite{wang2019rgb,wang2019learning,choi2020hi,wang2020cross} and feature-level alignment~\cite{ye2018visible,ye2018hierarchical,liu2020parameter,fu2021cm,wu2021discover}.
The image-level alignment-based methods aim to bridge the appearance discrepancy by transforming a visible (or infrared) image into its infrared (or visible) counterpart with Generative Adversarial Networks (GANs)~\cite{goodfellow2014generative}.
However, the image generation process usually suffers from high computational costs.
In addition, due to the lack of pose-aligned image pairs across modalities during training, GANs may inevitably introduce noisy generated samples~\cite{li2020infrared}, which leads to suboptimal performance.
The feature-level alignment-based methods typically design delicate network architectures~\cite{ye2018hierarchical,liu2020parameter,ye2020cross,wu2021discover} or exploit metric learning constraints~\cite{ye2021deep,liu2020parameter,ye2021channel} to learn a common feature space across modalities.
However, it is challenging to directly embed images of different modalities into a common feature space because of the significant color discrepancies between them.
Overall, previous methods primarily focus on learning modality invariant features through carefully designed network architectures, but few of them have analyzed which component of the image causes the modality discrepancy problem.
%

\begin{figure}[t]
\centering
\includegraphics[
width=0.99\columnwidth]{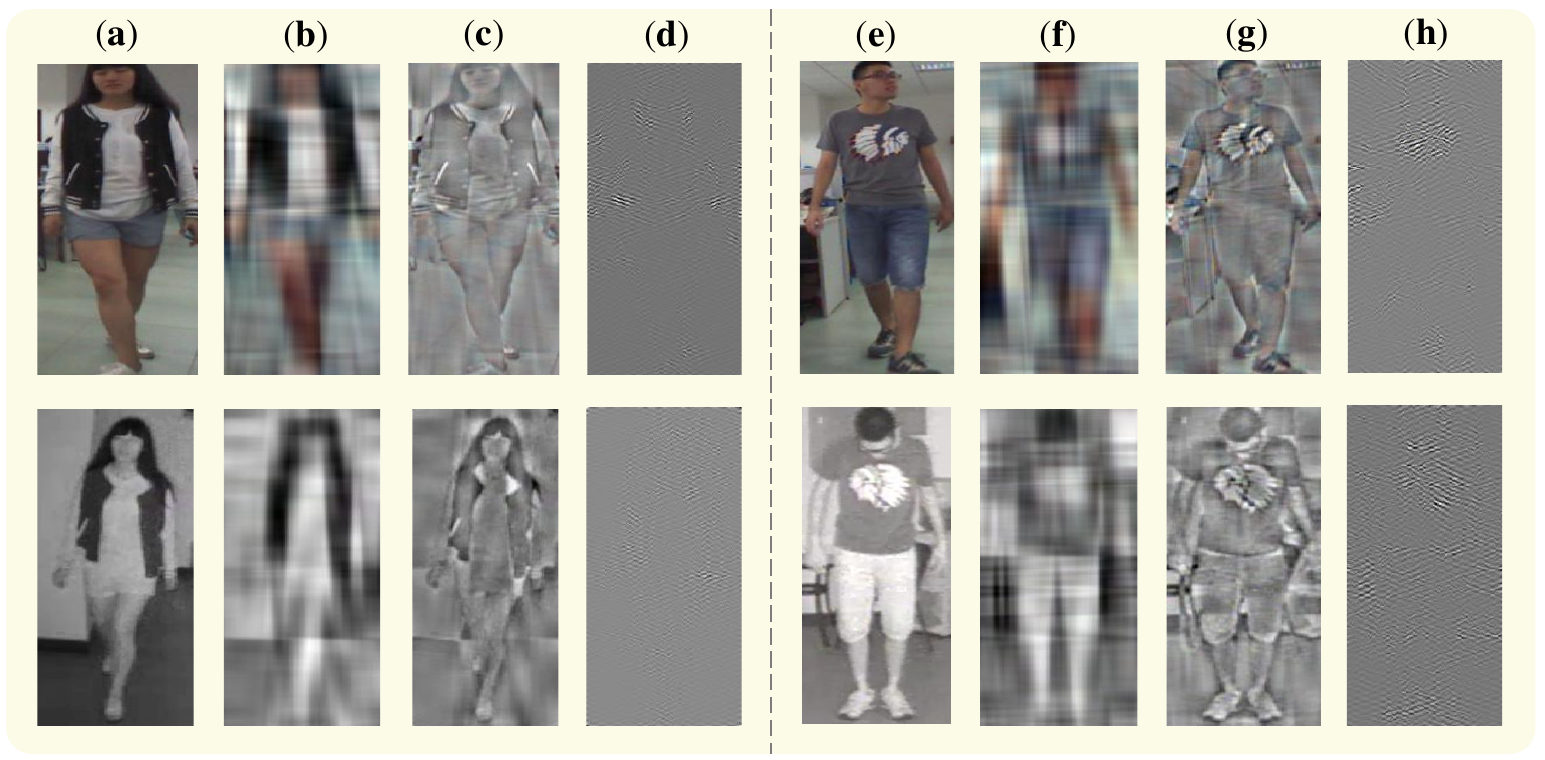}
\caption{
Examples of images reconstructed from filtered amplitude component.
(a) (e) Original images.
(b) (f) Reconstructed images from the low-pass filtered amplitude component and original phase component. 
(c) (g) Reconstructed images from the middle-pass filtered amplitude component and original phase component.
(d) (h) Reconstructed images from high-pass filtered amplitude component and original phase component.
It can be observed that different frequency components of the amplitude are of different transferability across modalities.
}
\label{motivation2}
\vspace{-5mm}
\end{figure}

In this paper, we attempt to answer this question and mitigate the modality discrepancy problem from the frequency domain perspective.
Our motivation stems from a widely recognized property of the Fourier transformation, where the phase component of the Fourier spectrum preserves high-level semantics of the original image, while the amplitude component contains low-level statistics~\cite{oppenheim1981importance,piotrowski1982demonstration}.
That is, the amplitude and phase components of the Fourier spectrum correspond to the style and semantic information within an image~\cite{chen2021amplitude,xu2021fourier,lee2023decompose}.
For better understanding, we first decompose the visible and infrared images into phrase and amplitude components and reconstruct the image from \textbf{only} phrase component and amplitude component, respectively.
From the results in Figure~\ref{motivation1}, we can observe that the amplitude-only reconstruction contains style information of an image, and the phase-only reconstruction reveals the visual structures of an image.
Since visible and infrared images can be viewed as two types of images with different styles, we argue that the difference in the amplitude component of visible and infrared images is the primary factor that causes the modality discrepancy problem.
To further substantiate our argument, we apply three different filters on the amplitude component and reconstruct the image from the filtered amplitude component and original phrase component.
From the results in Figure~\ref{motivation2}, we can observe that the reconstructed images with different amplitude components have distinct visual similarities.
The results reveal that different frequency components of the amplitude are of different transferability across modalities.
Consequently, we are motivated to eliminate the modality-specific amplitude component of an image to mitigate the modality discrepancy problem.

Based on the above analysis, we propose a novel Frequency Domain Modality-invariant feature learning framework (\textbf{FDMNet}) for VI-ReID, which consists of two modules: the Instance-adaptive Amplitude Filter (IAF) module and the Phrase-Preserving Normalization (PPNorm) module.
In the instance-adaptive amplitude filter module, we attempt to enhance the modality-invariant amplitude component of an image and suppress the modality-specific component to achieve \textbf{the image-level alignment}.
Specifically, we obtain an instance-adaptive attention map for each image and use it to filter out the modality-specific frequency component of the amplitude.
The filtered images from visible and infrared modalities are then used to confuse a modality discriminator, which makes the instance-adaptive amplitude filter module focus on the modality-invariant amplitude component.
In addition, we propose a novel Phrase-Preserving Normalization module to learn modality-invariant features to achieve \textbf{the feature-level alignment}.
Inspired by the success of Instance Normalization in reducing the discrepancy among instances~\cite{wu2021discover,zhang2023mrcn}, PPNorm composes the phase of the original feature and the amplitude of the post-normalized feature to obtain the modality-invariant feature.
Since the post-normalized feature contains less modality information, the modality discrepancy in the amplitude component of the reconstructed feature can be alleviated.
By jointly optimizing the Instance-adaptive Amplitude Filter module and the Phrase-Preserving Normalization module, our FDMNet can learn the modality-invariant features at both the image-level and feature-level from the frequency domain.
Additionally, our method can serve as a plug-and-play module with little computation cost.
Thus, we integrate the proposed IAF and PPNorm modules into two existing methods and achieve consistent performance improvements.

Our contributions are summarized as follows:
\begin{itemize}
\item
We introduce a novel Frequency Domain Modality-invariant feature learning framework (\textbf{FDMNet}) to learn the modality-invariant features at both the image-level and feature-level.
To the best of our knowledge, this is the first work that mitigates modality discrepancy from the perspective of frequency domain decomposition.
\item
We introduce the Instance-adaptive Amplitude Filter (IAF) module and the Phrase-Preserving Normalization (PPNorm) module to enhance the modality-invariant amplitude component and suppress the modality-specific component at both the image-level and feature-level.
\item
We conduct extensive experiments and show that our proposed method outperforms state-of-the-art methods on two standard VI-ReID benchmarks.
We demonstrate the effectiveness of our method on two previous VI-ReID methods and achieve consistent improvements.
\end{itemize}

\section{Related Work}
\label{sec:rw}
In this section, we briefly overview representative works related to ours, including visible-infrared person ReID and frequency domain modeling in deep learning.

\textbf{Visible-Infrared Person ReID.}
Different from traditional single-modality person ReID, VI-ReID aims to match visible and infrared person images captured by disjoint cameras~\cite{wu2017rgb}.
Current VI-ReID methods mainly focus on image-level alignment~\cite{wang2019rgb,wang2019learning,choi2020hi,wang2020cross} or feature-level alignment~\cite{ye2018visible,ye2018hierarchical,liu2020parameter,fu2021cm,wu2021discover} to reduce the modality discrepancy.
The image-level alignment-based methods aim to bridge the appearance discrepancy by transforming the image from one modality to the other with Generative Adversarial Networks (GANs).
%
AlignGAN~\cite{wang2019rgb} transfers stylistic properties of infrared images to their visible counterparts for pixel and feature alignment jointly.
cmGAN\textit{et al.}~\cite{wang2020cross} generates cross-modality paired images and performs global set-level and fine-grained instance-level alignments.
However, due to the lack of pose-aligned image pairs across modalities, it is difficult to preserve identity information for generated images~\cite{chen2021neural}.
As for the feature-level alignment, some studies exploit two-stream CNNs with deep metric learning~\cite{liu2020parameter,ye2020cross,ye2018visible,ye2020visible,liu2022learning} or the attention mechanism~\cite{wei2020co,ye2020dynamic,jiang2022cross,li2022visible,zhang2022fmcnet} to learn modality-invariant feature representations.
Fu \textit{et al.}~\cite{fu2021cm} and Chen \textit{et al.}~\cite{chen2021neural} exploit the optimal two-stream architecture by neural architecture search to alleviate the modality discrepancy.
Ye \textit{et al.}~\cite{ye2020dynamic} propose a dynamic dual-attentive aggregation learning method to mine both intra-modality part-level and cross-modality graph-level contextual cues.
Nevertheless, existing methods rarely investigate which specific components of the image contain modality-invariant information.
Differently, we attempt to explore which specific component of the image contains the modality-invariant information and alleviate modality discrepancy with the characteristics of images in the frequency domain.
%

%

\noindent\textbf{Frequency Domain in Deep Learning.}
Frequency analysis has been widely used in conventional digital image processing for decades.
Recently, information processing in the frequency domain has attracted increasing attention in deep learning~\cite{chi2019fast,xu2021fourier,wang2020high,rao2021global,xu2020learning}.
For example, Xu \textit{et al.}~\cite{xu2021fourier} find that the Fourier phase information contains high-level semantics and propose a Fourier-based augmentation strategy for domain generalization.
Rao \textit{et al.}~\cite{rao2021global} learn long-term spatial dependencies in the frequency domain to replace the self-attention layer in vision transformers.
Jiang \textit{et al.}~\cite{jiang2021focal} propose a novel focal frequency loss to directly optimize generative models in the frequency domain.
Cai \textit{et al.}~\cite{cai2021frequency} propose a  frequency domain image translation (FDIT) framework to exploit frequency information for enhancing the image generation process.
Motivated by the success of these works, we propose a novel FDMNet for VI-ReID, which is the first work to achieve image-level and feature-level alignment with frequency domain properties.

\section{Our Approach}
\label{sec:our}
\begin{figure*}
\centering
\includegraphics[width=0.90\linewidth]{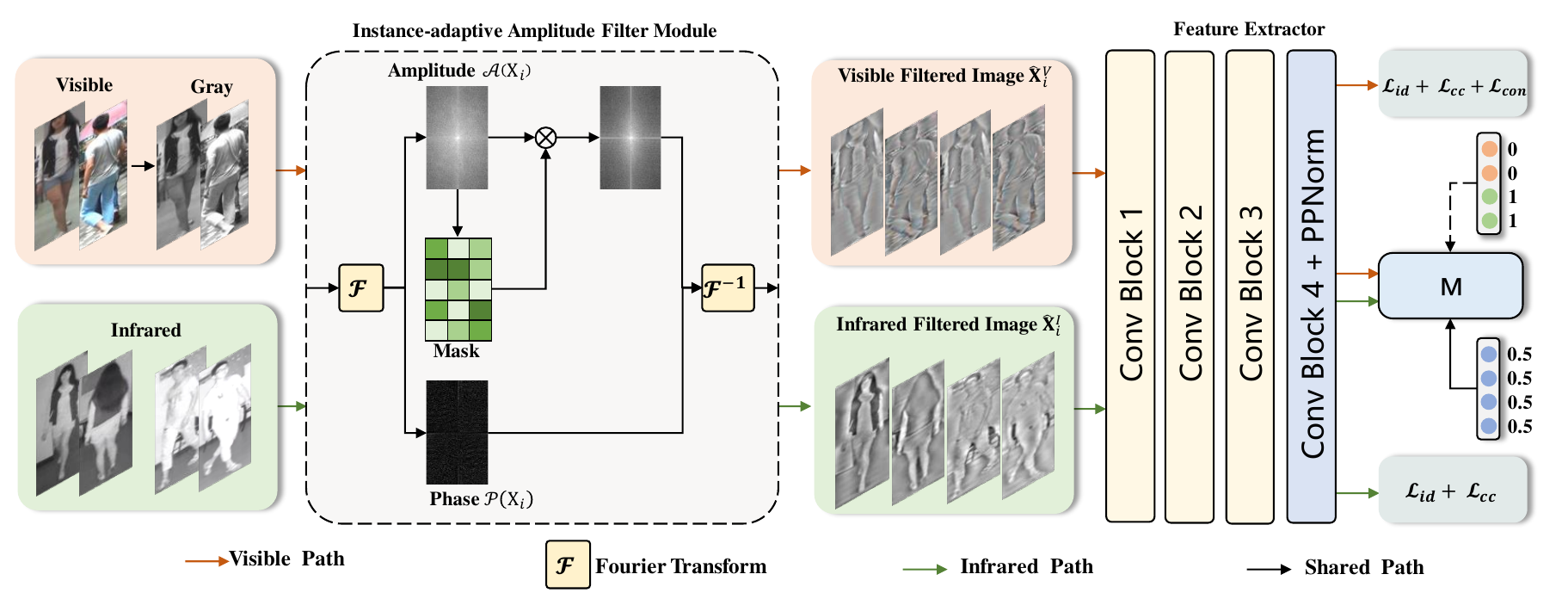}
\caption{
The overall architecture of our proposed Frequency Domain Modality-invariant feature learning framework (FDMNet).
Our FDMNet includes an Instance-adaptive Amplitude Filter (IAF) module and a Phrase-Preserving Normalization (PPNorm) module. The IAF and the PPNorm modules can enhance the modality-invariant amplitude component and suppress the modality-specific component at image-level and feature-level, respectively.
%
}\label{framework}
\end{figure*}

In this section, we first formulate the task of VI-ReID.
Then we describe each component of the proposed Frequency Domain Modality-invariant feature learning framework (FDMNet) in detail.

\subsection{Problem Formulation}\label{sec-31}
Let $\mathcal{V}=\left\{\mathbf{X}_i^{\boldsymbol{V}}\right\}_{i=1}^{N_V} \text { and } \mathcal{I}=\left\{\mathbf{X}_i^{\boldsymbol{I}}\right\}_{i=1}^{N_I}$ respectively denote the visible and infrared images, where ${N_V}$ and ${N_I}$ are the numbers of images in each modality.
$\mathcal{Y}^{\boldsymbol{V}}=\left\{{\mathbf{y}}_i^{\boldsymbol{V}}\right\}_{i=1}^{N_p}$ and $\mathcal{Y}^{\boldsymbol{I}}=\left\{{\mathbf{y}}_i^{\boldsymbol{I}}\right\}_{i=1}^{N_p}$ indicates the corresponding identity labels, where $N_p$ is the number of identities.
Given a query person image, the VI-ReID task aims to retrieve the same person by finding a ranked list of images from the other modality according to the feature similarity.

As shown in Figure~\ref{framework}, our FDMNet consists of three modules, including an Instance-adaptive Amplitude Filter (IAF) module, a modality-shared feature extractor, and a modality discriminator.
We first filter out the modality-specific amplitude component for the given image with the Instance-adaptive Amplitude Filter (IAF) module.
We also use grayscale images to assist visible modality learning with color-independent information.
To further mitigate the modality discrepancy at the feature level, we introduce the Phrase-Preserving Normalization (PPNorm) module into the feature extractor, which composes the phase of the original feature and the amplitude of the post-normalized feature to obtain the modality-invariant feature.
Finally, we employ a modality discriminator to minimize the difference between the feature distributions of visible and infrared modalities to learn modality-invariant features.
To better illustrate our approach, we distinguish visible and infrared modalities with $\boldsymbol{V}$ and $\boldsymbol{I}$ in superscript.

\subsection{Instance-adaptive Amplitude Filter}\label{sec-32}
As illustrated in the introduction, the amplitude and phase components of the Fourier spectrum correspond to the style and semantic information of an image.
Since visible and infrared images can be viewed as two types of images with different styles, we are inspired to filter out the modality-specific amplitude component of an image.

\textbf{Fourier transform.}
First, we revisit the operation of the Fourier transform.
Given a image $\mathbf{X} \in \mathbb{R}^{H \times W \times C}$, the Fourier transform $\mathcal{F}$ converts it to Fourier space as the complex component $\mathcal{F}(\mathbf{X})$, which is formulated as:
\begin{equation}
\mathcal{F}( \mathbf{X} )(u, v)=\frac{1}{\sqrt{H W}} \sum_{h=0}^{H-1} \sum_{w=0}^{W-1} \mathbf{X}(h, w) e^{-j 2 \pi\left(\frac{h}{H} u+\frac{w}{W} v\right)},
\end{equation}
where $j$ is the imaginary unit, and we denote $\mathcal{F}^{-1}( \mathbf{X} )$ as the inverse Fourier transform accordingly.
Both the Fourier transform and the inverse procedure can be efficiently implemented with the FFT algorithm~\cite{frigo1998fftw}.
In the Fourier space, each complex component $\mathbf{X}(u, v)$ ca be represented by the amplitude component $\mathcal{A}(\mathbf{X}(u, v))$ and the phrase component $\mathcal{P}(\mathbf{X}(u, v))$.
These two components are formulated as:
\begin{equation}
\begin{aligned}
& \mathcal{A}(\mathbf{X}(u, v))=\sqrt{R^2(\mathbf{X}(u, v))+I^2(\mathbf{X}(u, v))}, \\
& \mathcal{P}(\mathbf{X}(u, v))=\arctan \left[\frac{I(\mathbf{X}(u, v))}{R(\mathbf{X}(u, v))}\right],
\end{aligned}
\end{equation}
where $R(\mathbf{X}(u, v))$ and $I(\mathbf{X}(u, v))$ denote the real and imaginary parts of $\mathbf{X}(u, v)$.

\textbf{Amplitude filtering.}
Since the amplitude $\mathcal{A}(\mathbf{X})$ contains the style information of the image, we aim to filter out modality-specific amplitude components with instance-adaptive masks.
The Amplitude filtering process is the same for visible and infrared images. Here we take the visible image $\mathbf{X}_i^{\boldsymbol{V}}$ as an example.
We first obtain the amplitude component of $\mathbf{X}_i^{\boldsymbol{V}} \in \mathbb{R}^{H \times W \times C}$ as $\mathcal{A}(\mathbf{X}_i^{\boldsymbol{V}}) \in \mathbb{R}^{H \times\left(\left\lfloor\frac{W}{2}\right\rfloor+1\right) \times C}$.
Next, we adopt a $1 \times 1$ convolutional layer followed by Batch Normalization and ReLU activation to project $\mathcal{A}(\mathbf{X}_i^{\boldsymbol{V}})$ to a higher dimension as $\mathbf{A}_i^{\boldsymbol{V}}$.
We then apply global average pooling and global max pooling over channels on $\mathbf{A}_i^{\boldsymbol{V}}$ to obtain two frequency features.
Finally, we concatenate the frequency features along the channel dimension and use a $3 \times 3$ convolutional layer to learn the instance-adaptive amplitude mask.
Formally, 
\begin{equation}
\mathbf{M}_i^{\boldsymbol{V}} = \sigma\left(\operatorname{Conv}_{3 \times 3}\left(\left[\operatorname{AvgPool}\left( \mathbf{A}_i^{\boldsymbol{V}} \right), \operatorname{MaxPool} \left( \mathbf{A}_i^{\boldsymbol{V}} \right)\right]\right)\right),
\end{equation}
where $\sigma(\cdot)$ is the sigmoid function.
We multiply the instance-adaptive mask $\mathbf{M}_i^{\boldsymbol{V}}$ over $\mathcal{A}(\mathbf{X}_i^{\boldsymbol{V}})$ to filter out the modality-specific amplitude components:
\begin{equation}
\hat{\mathcal{A}}(\mathbf{X}_i^{\boldsymbol{V}}) = \mathbf{M}_i^{\boldsymbol{V}} \otimes \mathcal{A}(\mathbf{X}_i^{\boldsymbol{V}}),
\end{equation}
where $\otimes$ denotes the element-wise multiplication.
With the modulated amplitude component $\hat{\mathcal{A}}(\mathbf{X}_i^{\boldsymbol{V}})$, we can reconstruct the filtered visible image as:
\begin{equation}
\hat{\mathbf{X}}_i^{\boldsymbol{V}} = \mathcal{F}^{-1} ( \hat{\mathcal{A}}(\mathbf{X}_i^{\boldsymbol{V}}), \mathcal{P}(\mathbf{X}_i^{\boldsymbol{V}}) ).
\end{equation}
Our approach preserves the semantics of the reconstructed image by adding instance-adaptive masks exclusively to the amplitude components.
By filtering out the modality-specific amplitude components, we effectively mitigate the modality discrepancy. 
The same operation can be performed on infrared images to obtain the filtered image $\hat{\mathbf{X}}_i^{\boldsymbol{I}}$.

\textbf{Grayscale-guided Learning.}
Since significant color discrepancies exist between visible and infrared modalities, we adopt grayscale as guidance for IAF learning.
Specifically, we add a consistency loss to encourage the reconstructed images from IAF to be consistent between the visible image $\mathbf{X}_i^{\boldsymbol{V}}$ and its corresponding grayscale image $\mathbf{X}_i^{\boldsymbol{G}}$, since the color information is expected to be filtered out by the proposed IAF module.
Therefore, we add a consistency loss which is expressed as:
\begin{equation}\label{equ:recloss}
\resizebox{0.2\textwidth}{!}{$
\begin{split}
\mathcal{L}_{con} = \frac{1}{n} \sum_{i=1}^{n}  \left\| \hat{\mathbf{X}}_i^{\boldsymbol{V}} -  \hat{\mathbf{X}}_i^{\boldsymbol{G}}  \right\|_1 ,
\end{split}
$}\end{equation}
where $\hat{\mathbf{X}}_i^{\boldsymbol{G}}$ denote the grayscale reconstructed images from IAF.
The consistency loss can help the IAF module focus on color-irrelevant frequency components during training.

\subsection{Phrase-Preserving Normalization}\label{sec-33}
The proposed IAF module can remove modality-specific amplitude components from images to some extent.
However, directly applying it at the feature-level would result in high computational cost due to the high dimension of the feature map.
Recently, Instance Normalization has been proven to be effective in reducing the discrepancy among instances~\cite{wu2021discover,zhang2023mrcn}.
However, directly applying the IN layers would lead to losing some discriminative semantic information.
To address this issue, inspired by the fact that the phase components of the Fourier spectrum correspond to the semantic information, we propose a novel Phrase-Preserving Normalization that achieves semantic-consistent normalization at the feature-level.

\begin{figure}[t]
\centering
\includegraphics[
width=0.8\columnwidth]{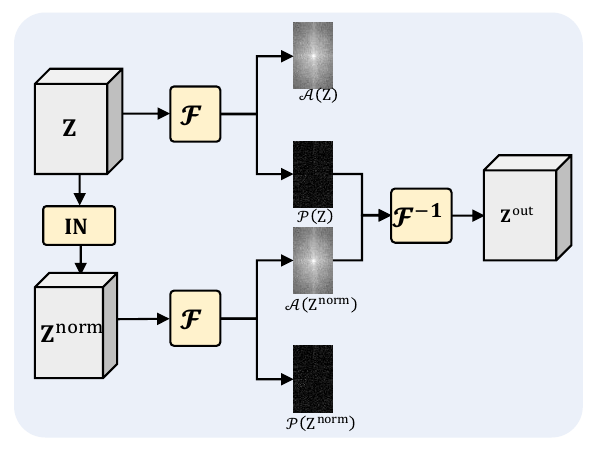}
\caption{
The detailed architecture of the proposed Phrase-Preserving Normalization (PPNorm) Module.
}
\label{AANorm}
\vspace{-5mm}
\end{figure}

Specifically, for an input image $\mathbf{X}_{i}$, we denote its feature map $\mathbf{Z} \in \mathcal{R} ^{h \times w \times c}$ extract by the convolutional block as the input of the PPNorm module, where $h,w$ and $c$ represent the height, width, and dimension of the feature map. 
The normalized feature can be obtained as follows:
\begin{equation}
{\mathbf{Z}}_k^{norm} = \operatorname{IN}\left(\mathbf{Z}_k\right)=\frac{\mathbf{Z}_k-\mathrm{E}\left[\mathbf{Z}_k\right]}{\sqrt{\operatorname{Var}\left[\mathbf{Z}_k\right]+\epsilon}}
\end{equation}
where $\mathbf{Z}_k$ denotes the $k$-th dimension of $\mathbf{Z}$, $\epsilon$ is used to avoid dividing-by-zero, $\mathrm{E}[\cdot]$ and $\operatorname{Var}[\cdot]$ represent the operation of calculating the mean and variance.
To maintain the semantic of the feature map $\mathbf{Z}$ and reduce the discrepancy among different modality features, we compose the phase of the original feature $\mathbf{Z}$ and the amplitude of the post-normalized feature $\mathbf{Z}^{norm}$ to obtain the modality-invariant feature.
Formally, the operation of PPNorm is formulated as follows:
\begin{equation}
\operatorname{PPNorm} \left( \mathbf{Z} \right) = \mathcal{F}^{-1}( \mathcal{A}( \mathbf{Z}^{norm} ), \mathcal{P}(\mathbf{Z})),
\end{equation}
where $\mathcal{F}^{-1}( \cdot )$ denote the inverse Fourier transform.
Since the phrase component of the output feature comes from the original feature $\mathbf{Z}$, the identity-related semantic information is preserved, while the phrase component from the post-normalized feature can help alleviate the modality discrepancy.
Figure~\ref{AANorm} illustrates the whole process of the proposed PPNorm.
We adopt the ResNet-50 as our feature extractor and replace the BN layers with the proposed PPNorm in stage-4 of ResNet-50 by default.

\subsection{Modality Adversarial Learning}\label{sec-34}
Similar to~\cite{hao2021cross}, we introduce the modality adversarial learning framework to reduce the discrepancy between visible and infrared modalities.
For the amplitude-filtered visible and infrared images, we denote the features extracted by the feature extractor as ${\mathbf{f}}_{i}^{\boldsymbol{V}}$ and ${\mathbf{f}}_{i}^{\boldsymbol{I}}$, respectively.
For each sample ${\mathbf{f}}_{i}$, we define a real modality label $\mathbf{t}_{i}$ and a fake modality label $\mathbf{o}_{i}$.
The real modality label is set to $[1,0]$ for the visible images and $[0,1]$ for the infrared images.
The fake modality label is set to $[0.5,0.5]$ for all samples from both modalities.
A modality discriminator M is adopted to distinguish the given image into a specific modality.
Specifically, for ${\mathbf{f}}_{i}$, the modality discriminator predict a two-dimensional probability $p_m \left( \mathbf{f}_{i} \right)$.
The modality discriminator is optimized to classify ${\mathbf{f}}_{i}$ into its real modality labels:
\begin{equation}
\mathcal{L}_m\left(\theta_m\right)=-\frac{1}{n+m} \sum_{i=1}^{n+m} \mathbf{t}_{i} \cdot \log p_m\left( \mathbf{f}_{i} \right),
\end{equation}
where $\theta_m$ denotes the parameters of the modality discriminator, $n$ and $m$ represents the number of visible and infrared images in the batch.
The feature extractor is optimized to confuse the modality discriminator to extract modality-invariant features:
\begin{equation}
\mathcal{L}_e\left(\theta_a, \theta_e\right)=-\frac{1}{n+m} \sum_{i=1}^{n+m} \mathbf{o}_{i} \cdot \log p_m\left( \mathbf{f}_{i} \right),
\end{equation}
where $\theta_a$ and $\theta_e$ denote the parameter of the Instance-adaptive Amplitude Filter module and the feature extractor, respectively.
During training, we update $\theta_m$ and $\theta_a$/$\theta_e$ alternately until they reach converge:
\begin{equation}
\begin{aligned}
& \mathcal{L}\left(\theta_m, \theta_e,\theta_a\right)=\mathcal{L}_m\left(\theta_m\right)+\mathcal{L}_e\left(\theta_e\right), \\
& \hat{\theta}_m=\arg \min _{\theta_m} \mathcal{L}\left(\theta_m, \hat{\theta}_e\right), \\
& \hat{\theta}_e=\arg \min _{\theta_a,\theta_e} \mathcal{L}\left(\hat{\theta}_a, \hat{\theta}_m, \theta_e\right) .
\end{aligned}
\end{equation}
Finally, the features extracted by the feature extractor cannot be correctly classified into their corresponding real modality labels, thereby achieving modality-invariant feature learning.
%
%

\subsection{Objective Functions}\label{sec-35}
With the features extracted by the feature extractor, we train the feature extractor and the IAF module with the identity loss and center-cluster loss~\cite{wu2021discover} to learn modality-invariant features.
The identity loss is formulated as follows:
\begin{equation}
\mathcal{L}_{id} = -\frac{1}{n} \sum_{i=1}^n {\mathbf{y}}_i^{\boldsymbol{V}} \log P \left( {\mathbf{f}}_{i}^{\boldsymbol{V}} \right) - \frac{1}{m} \sum_{i=1}^m {\mathbf{y}}_i^{\boldsymbol{I}} \log P \left(   {\mathbf{f}}_{i}^{\boldsymbol{I}} \right),
\end{equation}
where $n$ and $m$ denote the number of visible and infrared images in the batch, and $P ( \cdot )$ represents the identity prediction with a modality-shared classifier.
The center cluster loss is formulated as follows:
\begin{small}
\begin{equation}
\begin{aligned}
\mathcal{L}_{cc} & =\frac{1}{n+m} \sum_{i=1}^{n+m}\left\| {\mathbf{f}}_{i} -\mathbf{c}_{\mathbf{y}_i}\right\|_2 \\
& +\frac{2}{T(T-1)} \sum_{k=1}^{T-1} \sum_{j=k+1}^T \left[\rho-\left\|\mathbf{c}_{\mathbf{y}_i}-\mathbf{c}_{\mathbf{y}_j}\right\|_2\right]_{+},
\end{aligned}
\end{equation}
\end{small}
where $n$ and $m$ denote the numbers of visible images in the current batch, $\mathbf{c}_{\mathbf{y}_i}$ is the mean of modality-invariant features with label $\mathbf{y}_i$, $T$ is the number of identities in the current batch, and $\rho$ is the least margin among identity centers.
The center cluster loss aims to push the features into their center.
This loss builds the relationship among class centers rather than among samples, which can simultaneously alleviate the modality discrepancy and push away samples with different identities.

\subsection{Training and Inference}\label{sec-36}

The total loss of FDMNet is defined as:
\begin{equation}
\mathcal{L} = \mathcal{L}_{e}+ \mathcal{L}_{id} + {\lambda_1}\mathcal{L}_{con} + {\lambda_2} \mathcal{L}_{cc}, 
\end{equation}
where ${\lambda_1}$ and ${\lambda_2}$ are the hyper-parameter to balance the contribution of $\mathcal{L}_{con}$ and $\mathcal{L}_{cc}$.
During testing, the cross-modality matching is conducted by computing cosine similarities of feature vectors ${\mathbf{f}}_{i}^{\boldsymbol{V}}$ or ${\mathbf{f}}_{i}^{\boldsymbol{I}}$ between the probe and gallery images.


\section{Experiments}
\label{sec:expr}
\begin{table*}[t]
    \centering
    \caption{Comparisons with SOTA methods on the SYSU-MM01 and RegDB datasets.
       Rank-k ($\%$) and mAP ($\%$) are reported.}
    \label{sysu-mm01}
    \resizebox{1.0\textwidth}{!} {
    \begin{tabular}{l|c|ccc|ccc|ccc|ccc}
        \hline
        \multirow{3}{*}{Method} & \multirow{3}{*}{Venue} & \multicolumn{6}{c|}{SYSU-MM01} & \multicolumn{6}{c}{RegDB} \\
        \cline{3-14}
        & & \multicolumn{3}{c|}{All-Search} & \multicolumn{3}{c|}{Indoor-Search} & \multicolumn{3}{c|}{Visible to Infrared} & \multicolumn{3}{c}{Infrared to Visible}\\
        \cline{3-14}
        & & R-1 & R-10 & mAP & R-1 & R-10 & mAP & R-1 & R-10 & mAP & R-1 & R-10 & mAP \\
        \hline
        Zero-Pad~\cite{wu2017rgb} & ICCV-17 & 14.80 & 54.12  & 15.95 & 20.58 & 68.38  & 26.92 & 17.74 & 34.21 & 18.90 & 16.63 & 34.68  & 17.82 \\
        cmGAN~\cite{dai2018cross} & IJCAI-18 & 26.97 & 67.51 & 27.80 & 31.63 & 77.23 & 42.19 & - & - & - & - & - & - \\
        AlignGAN~\cite{wang2019rgb} & ICCV-19 & 42.4 & 85.0  & 40.7 & 45.9 & 87.6  & 54.3  & 57.9  & -  & 53.6  & 56.3  & -     & 53.4 \\
        Hi-CMD~\cite{wang2019rgb} & CVPR-20 & 34.94 & 77.85  & 35.94 & - & -  & - & 70.93 &86.39  & 66.04 & -   & -   & - \\
        DDAG~\cite{ye2020dynamic} & ECCV-20 & 54.75 & 90.39 & 53.02 & 61.02 & 94.06 & 67.98  & 69.34 & 86.19 & 63.46 & 68.06 & 85.15 & 61.80 \\
        NFS~\cite{chen2021neural} & CVPR-21 & 56.91 & 91.34 & 55.45 & 62.79 & 96.53 & 69.79  & 80.54 & 91.96 & 72.10 & 77.95 & 90.45 & 69.79 \\
        CM-NAS~\cite{fu2021cm} & ICCV-21 & 61.99 & 92.87 & 60.02 & 67.01 & 97.02 & 72.95  & 84.54 & 95.18 & 80.32 & 82.57 & 94.51 & 78.31 \\
        MCLNet~\cite{hao2021cross} & ICCV-21 & 65.40 & 93.33 & 61.98 & 72.56 & 96.98  & 76.58 & 80.31 & 92.7 & 73.07 & 75.93 & 90.93 & 69.49\\
        SMCL~\cite{wei2021syncretic} & ICCV-21 & 67.39 & 92.87 & 61.78  & 68.84 & 96.55 & 75.56 & 83.93 & -  & 79.83 & 83.05 & - & 78.57\\
        MPANet~\cite{wu2021discover} & CVPR-21 & 70.58 & 96.21 & 68.24 & 76.74 & 98.21  & 80.95 & 83.70 & -  & 80.90 & 82.80 & -  & 80.70 \\
        MID~\cite{huang2022modality} & AAAI-22 & 60.27 & 92.90  & 59.40 & 64.86 & 96.12  & 70.12 & 87.45 &95.73 & 84.85 & 84.29 &93.44 & 81.41 \\
        MAUM~\cite{liu2022learning} & CVPR-22 & 71.68 & - & 68.79 & 76.97 & - & 81.94 & 87.87 & -  & 85.09  & 86.95 & - & 84.34 \\
        CMT~\cite{jiang2022cross} & ECCV-22 & 71.88 & 96.45 & 68.57  & 76.90 & 97.68  & 79.91 & 95.17  & 98.82  & 87.30 & 91.97  & 97.92  & 84.46 \\
        DEEN~\cite{liu2022learning} & CVPR-23 & 74.7 & 97.6 & \bf{71.8} & 80.3 & \bf 99.0 & \bf{83.3} & 91.1 & 97.8  & 85.1   & 89.5 & 96.8 & 83.4 \\
        SGIEL~\cite{feng2023shape} & CVPR-23 & 75.18 & 96.87 & 70.12  & 78.40 & 97.46  & 81.20 & 91.07  & -  & 85.23 & 92.18  & - & 86.59 \\
        \hline 
        Our FDMNet & - & \bf{75.99} & \bf{97.63} & 70.71 & \bf{80.92} & 98.88 & 82.64  & \bf{95.92} & \bf{99.01}  & \bf{89.26} & \bf{93.58} & \bf{98.33} & \bf{86.88} \\
        \hline
    \end{tabular}}
\end{table*}

In this section, we first introduce the datasets and the implementation details.
Then, we verify the effectiveness of our method on two standard benchmarks and report a set of ablation studies to validate the effectiveness of each component.
Finally, we provide more visualization results.

\subsection{Datasets and Evaluation Protocol}\label{datasets}
\noindent\textbf{SYSU-MM01}~\cite{wu2017rgb} is the first large-scale benchmark for VI-ReID.
This dataset consists of 287,628 visible images taken by 4 visible cameras in the daytime, and 15,792 infrared images taken by 2 infrared cameras in the dark environment.
These images are captured in both indoor and outdoor scenarios.
The training set and the testing set have 395 and 96 identities, respectively.
There are two testing modes: \emph{all-search} and \emph{indoor-search}.
For the \emph{all-search} mode, the gallery set contains visible images in both indoor and outdoor scenarios.
For the \emph{indoor-search} mode, the gallery set merely contains visible images in the indoor scenario.

\noindent\textbf{RegDB}~\cite{nguyen2017person} is collected by a dual-camera system, including one visible and one thermal-infrared camera. 
There are 412 persons, and each person has 10 visible images and 10 infrared images.
Following~\cite{ye2018hierarchical}, 2,060 images from 206 person identities are randomly chosen as the training set and the remaining 2,060 images from 206 identities make up the testing set.
There are two evaluation settings: \emph{Visible to Infrared} and \emph{Infrared to Visible}. 
%
The former retrieves infrared images from visible ones, and the latter retrieves visible images from infrared ones. 

\noindent\textbf{Evaluation Metrics.}
Following standard evaluation protocols for VI-ReID~\cite{wu2017rgb,ye2021deep}, we adopt Cumulative Matching Characteristic (CMC) and mean Average Precision (mAP) for performance evaluation.
The reported result on the SYSU-MM01 dataset is an average performance of 10 times repeated random probe/gallery splits~\cite{wu2017rgb}, while that on the RegDB dataset is an average performance of 10 trials with different splits of training/testing sets~\cite{ye2018hierarchical,wang2019learning}.

\subsection{Implementation Details}\label{implement}
Following the previous VI-ReID methods~\cite{liu2022learning,zhang2022fmcnet,wu2021discover}, We adopt an ImageNet pretrained ResNet-50~\cite{he2016deep} as our backbone.
Following~\cite{liu2020parameter,ye2021deep,ye2020dynamic}, we set the stride of the last convolutional block as 1.
We resize each image to the size of 288 $\times$ 144 as in~\cite{jiang2022cross} and apply horizontal flipping and random erasing~\cite{zhong2020random} for data augmentation.
We randomly chose 8 identities for a mini-batch, each with 4 visible and 4 infrared images.
We use the SGD optimizer to train our model for 100 epochs with a batch size of 64.
The learning rate is initialized to $0.1$ and decayed to its $0.1$ and $0.01$ at the 20-th and 50-th epochs.
We implement our model with PyTorch and train it on one NVIDIA Telsa V100 GPU.

\subsection{Comparison with State-of-the-art Methods}\label{SOTA}

In this section, we compare the proposed FDMNet with the state-of-the-art VI-ReID methods on the SYSU-MM01 and  RegDB datasets.
The compared methods include image-level alignment-based methods (cmGAN~\cite{dai2018cross}, AlignGAN~\cite{wang2019rgb}), and  feature-level alignment-based methods (NFS~\cite{chen2021neural}, CM-NAS~\cite{fu2021cm}, MCLNet~\cite{hao2021cross}, SMCL~\cite{wei2021syncretic}, MID~\cite{huang2022modality}, MAUM~\cite{liu2022learning},  FMCNet~\cite{zhang2022fmcnet}, MPANet~\cite{wu2021discover},DCM~\cite{sun2022not},CMT~\cite{jiang2022cross},DEEN ~\cite{liu2022learning},SGIEL~\cite{jiang2022cross}).
We directly use the original results from published papers for comparison.

\noindent\textbf{Results on the SYSU-MM01 Dataset.}
As shown in table~\ref{sysu-mm01}, the proposed FDMNet achieves the best performance compared with the state-of-the-art under global and part feature settings.
Specifically, the FDMNet  achieves 75.99$\%$ Rank-1 accuracy and 70.71$\%$ mAP score in the most challenging \emph{all search} mode, improving the Rank-1 accuracy by 0.81$\%$ over the previous method SGIEL~\cite{jiang2022cross}, which verifies the superiority of the proposed frequency domain modality-invariant feature learning at both image-level and feature-level.
Further, compared with the image-level alignment-based methods, our FDMNet achieves much better performance and exceeds AlignGAN~\cite{wang2019rgb} by 33.59$\%$ in Rank-1 accuracy and 30.01$\%$ in mAP in the \emph{all search} mode.
This is because their performance is highly dependent on the quality of the generated images.
However, due to the lack of pose-aligned image pairs across modalities, it is challenging to generate high-quality images.
In contrast, our FDMNet can suppress the modality-specific frequency component and provide more discriminative representations.

\noindent\textbf{Results on the RegDB Dataset.}
The comparison results on the RegDB Dataset are shown in Table~\ref{sysu-mm01}.
The performance of FDMNet outperforms existing state-of-the-art methods under both evaluation modes.
Especially, in the \emph{visible-to-infrared} mode, FDMNet makes an improvement of 0.75$\%$ in Rank-1 accuracy and 1.96$\%$ in mAP compared to the top-performing method CMT~\cite{liu2020parameter}.
Our Rank-1 accuracy reaches 95.92$\%$, which is a marked high performance.
A similar improvement is also present in the \emph{infrared-to-visible} mode, which shows that our method is robust to visible and infrared query settings.
In conclusion, the results verify the effectiveness of the proposed frequency domain modality-invariant feature learning.

\subsection{Ablation Studies}\label{ablation}
In this section, we perform ablation studies on the SYSU-MM01 dataset in the \emph{all-search} mode to analyze each component of the proposed FDMNet with global features, including the IAF module, the Grayscale-guided learning strategy (${\mathcal{L}}_{con}$), the PPNorm module, and the Modality Adversarial Learning mechanism (MAL).
The results are shown in Table~\ref{tab:ablation}.

\noindent\textbf{Effectiveness of the Instance-adaptive Amplitude Filter.}
From index-1 and index-2, the performance with the IAF module is significantly improved compared to the baseline method, achieving a Rank-1 accuracy of $68.56\%$ and mAP of $65.39\%$.
This is because the difference in the amplitude component of visible and infrared images is the primary factor that causes the modality discrepancy.
The proposed IAF module enhances the modality-invariant amplitude component of an image while suppressing the modality-specific component, leading to a significant improvement.

\begin{table}[]
\centering
\small
\caption{Performance comparison with different components of our method on the \textbf{SYSU-MM01} dataset in the \emph{all-search} mode.}
\label{tab:ablation}
\resizebox{0.475\textwidth}{!} {
\begin{tabular}{c|cccc|ccc|c}
\myline
\multirow{2}{*}{Index} &\multirow{2}{*}{IAF}    &  \multirow{2}{*}{${\mathcal{L}}_{con}$} &  \multirow{2}{*}{PPNorm}  &  \multirow{2}{*}{MAL} & \multicolumn{4}{c}{SYSU-MM01}  \\
\cline{6-9}
& & & & &   R-1 &R-10   &R-20  &mAP\\
\hline
1   &   &           &           &                       &65.14     &93.54     &95.42     &62.56   \\
2   &$\checkmark$    &     &     &                      &68.56     &94.24     &96.98     &65.39  \\
3   &$\checkmark$    &$\checkmark$    &    &            &69.82     &95.60     &97.72     &67.74   \\
4   &    &    &$\checkmark$              &              &68.93     &95.63     &97.14     &66.21   \\
5   &$\checkmark$    &$\checkmark$    &$\checkmark$ &   &73.86     &96.75     &98.34     &69.06   \\
6   &$\checkmark$   &$\checkmark$    &$\checkmark$   &$\checkmark$   &\textbf{75.99}   &\textbf{97.63}     &\textbf{98.94}    &\textbf{70.71}   \\
\myline
\end{tabular}}
\end{table}

\noindent\textbf{Effectiveness of the Grayscale-guided Learning Strategy.}
From index-2 and index-3, we can see that when $\mathcal{L}_{o}$ is added, the performance is improved by $+1.26\%$ Rank-1 accuracy and  $+2.35\%$ mAP.
This is because The proposed ${\mathcal{L}}_{con}$ encourages the reconstructed images from IAF to be consistent between the visible image and its corresponding grayscale image, such that can help the IAF module focus on color-irrelevant frequency components during training.

\noindent\textbf{Effectiveness of the Phrase-Preserving Normalization.}
From index-1 and index-4, we can see that when PPNorm is added into the feature extractor, the performance is improved by $+3.79\%$ Rank-1 accuracy and  $+3.65\%$ mAP.
This is because the semantic-related phrase component is preserved by PPNorm, while the phrase component from the post-normalized feature can help to alleviate the modality discrepancy problem.
From index-3 and index-5, we can observe that when incorporating the IAF module and PPNorm module together,
the performance is further improved by $+4.04\%$ Rank-1 accuracy, which shows that these two modules are complementary and can promote each other.

\noindent\textbf{Effectiveness of the Modality Adversarial Learning.}
From index-5 and index-6, we can see that when the Modality Adversarial Learning is added, the performance is improved by $+1.65\%$ and up to $+70.71\%$ mAP.
These results demonstrate the effectiveness of the modality adversarial learning mechanism, which can make the embeddings extracted by the feature extractor incorrectly classified into the corresponding modality, achieving modality-invariant feature learning framework.

\noindent\textbf{Impact of Different Filtering Strategies.}
To verify the effectiveness of the proposed Instance-adaptive Amplitude Filtering module, we conduct experiments to explore the impact of different filtering strategies.
(1) Modality-adaptive Spatial Filtering (MSF), which learns a spatial mask for each modality and use it to filter the image in the spatial domain.
(2) Instance-adaptive Spatial Filtering (ISF), which generates a spatial mask for each instance and use it to filter the image in the spatial domain.
(3) Modality-adaptive Amplitude Filtering (MAF), which learns a mask for each modality and use it to filter the amplitude component of the image.
The experimental results are shown in Table~\ref{tab:filter}.
The results indicate that the learnable mask in the spatial domain may have a negative impact since it is difficult to learn a mask directly from the image.
In addition, the proposed IAF also outperforms the MAF, which shows that we need to filter the amplitude component based on the instance-adaptive information.

\begin{table}[]
\centering
\small
\caption{Effectiveness of different filtering strategies on the SYSU-MM01 dataset in the \emph{all-search} mode.
MSF refers to Modality-adaptive Spatial Filtering, ISF represents Instance-adaptive Spatial Filtering, and MAF refers to Modality-adaptive Amplitude Filtering.
}
\label{tab:filter}
\resizebox{0.34\textwidth}{!} {
\begin{tabular}{c|ccc|c}
\myline
\multirow{2}{*}{Method} & \multicolumn{4}{c}{SYSU-MM01}   \\
\cline{2-5}
                &R-1       & R-10        &R-20           &mAP     \\
\hline
MSF             &69.12      &95.74        &97.66         &65.37              \\
ISF             &66.62      &95.11        &96.24         &65.02              \\
MAF             &71.28      &96.55        &98.01         &68.88                \\
IAF (Ours)      &75.99      &97.63        &98.94       &70.71      \\
\myline
\end{tabular}}
\end{table}

\noindent\textbf{Impact of Different Normalization Layers.}
To further verify the effectiveness of our proposed normalization layer, we integrate the proposed PPNorm and Instance Normalization into the baseline model, and the experimental results are shown in Table~\ref{tab:concat}. 
We insert them in stage-4 of the ResNet-50, respectively.
It can be seen that directly applying IN can improve performance. 
In addition, our proposed PPNorm can alleviate the modality discrepancy as well as maintain the semantic information of the image, thus achieving the best performance.

\begin{table}[]
\centering
\small
\caption{Effectiveness of different Normalization layers on the SYSU-MM01 dataset in the \emph{all-search} mode.}
\label{tab:concat}
\resizebox{0.4\textwidth}{!} {
\begin{tabular}{c|ccc|c}
\myline
\multirow{2}{*}{Method} & \multicolumn{4}{c}{SYSU-MM01}   \\
\cline{2-5}
                              &R-1       & R-10        &R-20           &mAP     \\
\hline
Base                         &65.14      &93.54        &95.42         &62.56              \\
Instance Norm       &67.49      &94.11        &96.97         &64.87                \\
Ours                         &68.93      &95.63        &97.14       &66.21   \\
\myline
\end{tabular}}
\end{table}

\subsection{Further Analysis}\label{model_analysis}

\noindent\textbf{Generalization of our framework}.
The proposed IAF and PPNorm modules can serve as plug-and-play modules to existing methods with little computation cost.
To demonstrate the generalization of our proposed frequency domain modality-invariant feature learning, we incorporated the IAF and PPNorm modules into two feature-level alignment-based methods, AGW~\cite{ye2021deep} and HCT~\cite{liu2020parameter}.
For AGW~\cite{ye2021deep}, we reproduce it with additional random erasing data augmentation for better performance.
The experimental results are shown in Table~\ref{tab:generalization}.
From the results, we can see that performance results are consistently improved when IAF and PPNorm modules are incorporated.
For example, on the SYSU-MM01 dataset, our framework improves AGW with $+4.58\%$ Rank-1 accuracy and $+2.65\%$ mAP, respectively.
The results indicate that our method has strong generalization capability and can be integrated into existing models to improve performance.

\begin{table}[]
\centering
\small
\caption{Effectiveness of the proposed IAF and PPNorm modules over different methods on the SYSU-MM01 and RegDB datasets.}
\label{tab:generalization}
\resizebox{0.44\textwidth}{!} {
\begin{tabular}{c|cc|cc|c}
\myline
\multirow{2}{*}{Method} & \multicolumn{2}{c|}{SYSU-MM01}   &  \multicolumn{2}{c|}{RegDB} & \multirow{2}{*}{\#Params(M)}   \\
\cline{2-5}
                              &R-1       & mAP       &R-1           &mAP &    \\
\hline
AGW~\cite{ye2021deep}         &58.32      &58.58       &79.51      &73.52    &23.55             \\
AGW+Ours                      &62.90      &61.23       &82.35      &76.04    &24.02              \\
HCT~\cite{liu2020parameter}   &61.68      &57.51       &91.05      &83.28    &27.49             \\
HCT+Ours                      &66.42      &65.12       &92.98      &86.48    &27.95              \\
\myline
\end{tabular}}
\end{table}

\section{Conclusions}
\label{sec:conclus}
In this paper, we propose a novel Frequency Domain Modality-invariant feature learning framework  (FDMNet) to provide a new frequency domain perspective for the VI-ReID task.
We introduce the Instance-adaptive Amplitude Filter (IAF) module and the Phrase-Preserving Normalization (PPNorm) module to enhance the modality-invariant amplitude component and suppress the modality-specific component at both the image- and feature-levels.
Extensive experiments on two standard benchmarks demonstrate the superiority of the proposed FDMNet.
%
%
We hope that our work can inspire more visible-infrared applications and further investigate frequency domain properties.

{
    \small
    \bibliographystyle{ieeenat_fullname}
    \bibliography{cvpr_ref}
}


\end{document}